# Digital Overconsumption and Waste: A Closer Look at the Impacts of Generative AI

Vanessa Utz
Simon Fraser University
Burnaby, BC  V5A 1S6
`vutz@sfu.ca`

Steve DiPaola
Simon Fraser University
Burnaby, BC  V5A 1S6
`sdipaola@sfu.ca`

## Abstract

*Generative Artificial Intelligence (AI) systems currently contribute negatively to the production of digital waste, via the associated energy consumption and the related CO2 emissions. At this moment, a discussion is urgently needed on the replication of harmful consumer behavior, such as overconsumption, in the digital space. We outline our previous work on the climate implications of commercially available generative AI systems and the sentiment of generative AI users when confronted with AI-related climate research. We expand on this work via a discussion of digital overconsumption and waste, other related societal impacts, and a possible solution pathway.*

## 1. Introduction

Over the last year, commercially available generative AI systems, such as Midjourney and Stable Diffusion, which allow users to produce large quantities of digital artworks within increasingly short timespans, have attracted large user bases. Midjourney's Discord server, through which the system operates, currently has over 12 million members, while StabilityAI confirmed in October 2022, that Stable Diffusion had over 10 million daily users [1]. Estimates have put the daily global output of these systems at over 20 million images [2]. With this widespread adoption of these commercially available generative AI systems, there is a need to investigate the broader implications of these systems. In previous work [3], we provided preliminary estimates on the annual global energy consumption associated with the use of these systems. Our estimates were in the range of 1.92 TWh and 9.29 TWh, equivalent to the annual electricity consumption of Mauritania and Kenya respectively. The large range between the low and high estimates is due to the fact that exact daily user numbers, hours of use, as well as information on utilized hardware, are not readily available to researchers. In subsequent work [4] (still under review), we investigate the sentiment of generative AI users towards AI-related climate research. The motivation behind this work was to gain insight into the motivation behind generative AI use and whether users ever contemplate potential climate implications associated with their use of generative AI. We also used this study to check whether our assumptions underlying our initial energy consumption estimation in [3] were correct. Our results showed that while climate change was categorized as a "very urgent" issue by the majority of our study participants, most had not considered the energy consumption associated with generative AI. Furthermore, open-ended responses provided by participants after they were presented with data on the C02 emission associated with training Machine Learning (ML) models [5] and our own energy consumption estimates, indicated that a large number of participants were either skeptical of the data provided or pessimistic about how other users would respond to this data. The data does not suggest that users within the generative AI community would be open to changes in behavior based on the knowledge of the associated environmental impacts.

In this workshop, we want to present our on-going work on the impact of AI on our environment and society. We present a brief discussion on digital overconsumption and waste, how generative AI systems currently contribute to this problem, a possible solution pathway as well as wider societal implications related to widespread adoption generative AI. The goal is to spark much needed conversations around these issues during this rapid development of generative AI.

## 2. Digital Overconsumption and Generative AI

The concept of digital waste is important to consider in this discussion on generative AI and its climate implications. Digital waste (or data waste) is defined as the emissions and waste, as well as the extraction of natural resources and other destructive environmental practices related to the creation, use and maintenance of data infrastructures [6]. Our data showed an average weekly number of over 1500 generated images per users [4]. Additionally, a brief survey that we conducted as part of our initial work on energy consumption of generative AI [3], indicated that around 40% of respondents use the tools solely for themselves as a manner of entertainment.

Approximately half of our respondents also indicated that they require over 50 iterations on an idea to achieve a satisfying result. These numbers show that a significant portion of the generated images are not created and used for a direct utilitarian purpose. Most images that are created using these tools are currently never looked at again after the initial creation. It should also be pointed out that our data shows that less than 15% of our respondents classify themselves as "professional users" (participants were classified as professional users if they used generative AI predominately as part of their professional creative practice, i.e. as an artist or designer). The user base of generative AI systems appears to mainly consist of casual users who currently generate a lot of output for non-utilitarian purposes.

As we already outlined in more detail elsewhere [3], we hypothesize that one possible explanation for the recent explosion of use of generative AI, and the high number of image output per user that we are observing, is related to uses and gratification theory [7], a widely cited mass communication framework to study how media can satisfy a person's needs and desires, leading to the continued and prolonged consumption. It has already been applied in the investigation into the seemingly addictive nature of social media platforms such as TikTok [8]. Other work has shown that a need for escapism has been linked to increased consumption of digital content amongst individuals [9]. Generative AI systems now allow users to create artworks at a quality that would have up until now required high levels of artistic skill and a significant time investment (both to acquire the skills and then to execute the artwork), which could be an important factor as to why generative AI has become a highly attractive content generation mechanism. Future work is required to investigate whether gratification theory also applies in the context of generative AI, which now allows an almost instant creation and consumption of digital content, which would not be possible otherwise.

## 3. Other Societal Impacts & Future Work

Other societal implications are linked to the rapid development of generative AI and its incorporation in other existing technologies. While related to the idea of mere digital overconsumption in the present, these issues are most likely going to take shape in the future. They are however still important to discuss at this point in time, due to the very fast developments in the area of generative AI. Early identification of potential implications can lead to the faster development of solutions, such as new tools, best practices etc.

We posit that further research of how generative AI systems interact with our cognitive reward system is especially needed here as generative AI systems develop from mainly generating still imagery to the creation full video and immersive environments within 3D virtual reality (VR) systems. We have hypothesized in previous work that there could be potential negative societal impacts related to the correlation of heightened need for escapism and the newfound ability of users to now instantly create artificial worlds that could serve as virtual sanctuaries from a reality that is perceived as undesirable [3]. Potential dangers could lie amongst the increased social isolation. However, these sanctuaries do not only have potential negative societal implications. Future work should also investigate the AI for social good avenues that present themselves through the application of generative AI within 3D VR worlds.

Due to the widespread adoption of these tools, such work could also provide valuable insights into everyday human-AI/human-technology interaction and meaning making in the digital space. This could further our understanding of the societal impact of these new technologies.

## 4. Possible Solutions

There is no unilateral solution to this problem concerning the digital overconsumption associated with generative AI. Here, we merely want to outline one possible solution pathway centered around education. Based on the responses in our survey [3] and study [4], we believe that the need for a new education approach on the topic is high. Participants voiced their opinion on this strongly in our study [4]. Many did not believe that change in user behavior will come about without an education campaign within AI art communities, or potential legislative intervention into how these systems are developed and deployed. Future work needs to investigate more avenues to address this problem, such as hardware design and model architecture to reduce the energy draw of these systems.

We want to begin the academic discussion around this solution by using a framework [10] that has categorized overconsumption as a form of the common pool source dilemma, where 1) the full size of the resource pool is not known, 2) access to the resources in not equally distributed among individuals and 3) individuals must make decision on their consumption of goods and services without a full picture of the quantities and types of resources required in the process. We believe in the application of this common pool source dilemma to the issue of overconsumption and generative AI, as we have demonstrated in our previous work that a significant number of users are unaware of their own, never mind the total global, energy consumption (and the associated greenhouse gas (GHG) emissions) involved in running generative AI systems. Users are therefore unable to make informed decisions regarding their own behavior. While an individual running Stable Diffusion on their home computer might not impact global energy consumption, the impact is substantially magnified, at the scale at which these systems are currently being utilized globally. Ideally, future cross-disciplinary approaches can lead to solutions on how to raise awareness among the different stakeholders.